\documentclass[twocolumn]{article}

\usepackage{graphicx}

\newcommand{\bi}[1]{{\bf #1}}
\newcommand{\inner}[2]{\langle#1,#2\rangle}

\begin{document}

\title{
A kernel method for canonical correlation analysis 
\thanks{This is the full version of paper
presented in IMPS2001 (International Meeting
of Psychometric Society, Osaka, 2001)
}
}

\author{
Shotaro Akaho \\
AIST Neuroscience Research Institute, \\
Central 2, 1-1 Umezono, Tsukuba, Ibaraki 3058568, Japan \\
s.akaho@aist.go.jp \\
http://staff.aist.go.jp/s.akaho/
}
\date{}
\maketitle

\begin{abstract}
Canonical correlation analysis is a technique to extract common
features from a pair of multivariate data. In complex situations,
however, it does not extract useful features because of its linearity.
On the other hand, kernel method used in support vector
machine is an efficient approach to improve such a linear method.
In this paper, we investigate the effectiveness of applying kernel method
to canonical correlation analysis. 
\end{abstract}

\section{Introduction}
This paper deals with the method to extract common features from
multiple information sources.
For instance, let us consider a task of learning in pattern recognition,
in which an object is given by using an image and its name is
given by a speech.
For a newly given image, the system is required to answer its name by
a speech, and for a newly given speech, the system is to answer the
corresponding image.
The task can be considered to be a regression problem from image to speech
and vice versa. However, since the dimensionalities of images and speeches
are generally very large, a regression analysis many not work effectively.
In order to solve the problem, it is useful to map the inputs into
low dimensional feature space and then to solve the regression problem.

The canonical correlation analysis (CCA) has been
used for such a purpose. CCA finds a linear transformation of a pair of
multi-variates such that the correlation coefficient is maximized.
From an information theoretical point of view,
the transformation maximizes the mutual information between extracted
features. However, if there is nonlinear relation between the 
variates, CCA does not always extract useful features.

On the other hand, the support vector machines (SVM) are attracted
a lot of attention by its state-of-art performance in pattern recognition
\cite{vapnik}．
The kernel trick used in SVM is applicable not only for classification
but also for other linear techniques, for example,
kernel regression and kernel PCA\cite{scholkopf}． 

In this paper, we apply the kernel method to CCA.
Since the kernel method is likely to overfit the data,
we incorporate some regularization technique to avoid the overfitting.

\section{Canonical correlation analysis}

\begin{figure}[htbp]
  \begin{center}
   \includegraphics[width=.4\textwidth]{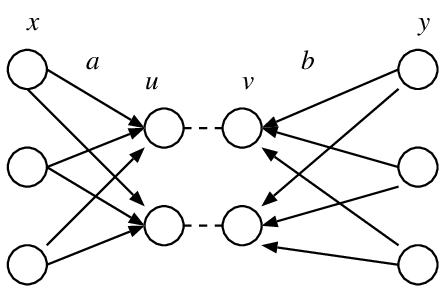}
    \caption{CCA}
    \label{fig:cca}
  \end{center}
\end{figure}

CCA has been proposed by Hotelling in 1935\cite{anderson}.
Suppose there is a pair of multi-variates
$\bi{x}\in {\cal R}^{n_x}$ と $\bi{y}\in {\cal R}^{n_y}$,
CCA finds a pair of linear transformations
such that the correlation coefficient between extracted features
is maximized
(Fig.\ref{fig:cca})．
For the sake of simplicity,
we assume that the averages of $\bi{x}$ and $\bi{y}$ are {\bf 0},
and the dimensionality of feature is 1, then by the transformations
\begin{equation}
  u = \inner{\bi{a}}{\bi{x}}, 
\end{equation}
\begin{equation}
  v = \inner{\bi{b}}{\bi{y}},
\end{equation}
we would like to find the transformation $\bi{a}$, $\bi{b}$ that maximizes
\begin{equation}
  \rho = {{\rm E}[u v]\over \sqrt{{\rm Var}[u] {\rm Var}[v]}},
\end{equation}
where $\inner{\bi{a}}{\bi{x}}$ represents the inner product.
We have to further assume
\begin{equation}
  {\rm Var}[u] = {\rm Var}[v] = 1,
\end{equation}
to reduce the freedom of scaling of $u$ and $v$.
$\bi{a}$ and $\bi{b}$ can be found by an eigen vector corresponding
to the maximal eigen values of a generalized
eigen value  problem.
If we need more than one dimension, we can take eigen vectors corresponding
other maximal eigen values.

CCA is important in an information theoretical viewpoint, since
it finds a transformation that maximizes the mutual information between
features, when $\bi{x}$ and $\bi{y}$ are jointly Gaussian.
Even if the assumption is not fullfilled, CCA can be still used in some
cases. However, if the purpose is regression, the large values of
correlation coefficients are crucially necessary.
The reasons that correlation coefficients are small can be
considered in the following cases:
\begin{enumerate}
\item 
  $\bi{x}$ and $\bi{y}$ does not have almost any relation.
\item There is strong nonlinear relation between $\bi{x}$ and $\bi{y}$,
\end{enumerate}
It is impossible of improvement in the first case.
However, in the second case, we can obtain the relation by some
methods.
One of those methods is to allow the nonlinear transformation and 
Asoh et al\cite{asoh} has proposed a neural network model that approximates
the optimal nonlinear canonical correlation analysis.
However, this model requires a lot of computation time and
it also has a lot of local optima.
In this paper, we incorporate the kernel method, which enables the
nonlinear transformation as well as the small computation and
no undesired local optima.

\section{Kernel CCA}

\begin{figure}[htbp]
  \begin{center}
    \includegraphics[width=.4\textwidth]{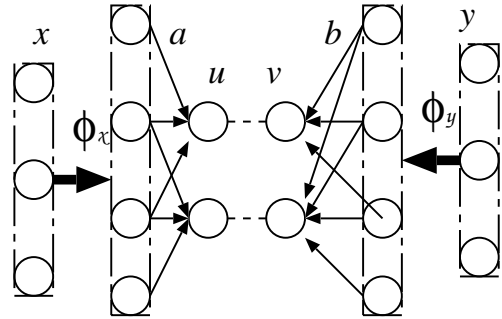}
    \caption{Kernel CCA}
    \label{fig:kcca}
  \end{center}
\end{figure}

First, $\bi{x}$ and $\bi{y}$ are transformed into the Hilbert space,
$\phi_x(\bi{x})\in H_x$ and $\phi_y(\bi{y})\in H_y$.
By taking inner products with a parameter in the Hilbert spaces,
$a\in H_x$, and $b\in H_y$, we find a feature
\begin{equation}
  u = \inner{a}{\phi_x(\bi{x})}, 
\end{equation}
\begin{equation}
  v = \inner{b}{\phi_y(\bi{y})},  
\end{equation}
which maximizes the correlation coefficients.

Now, suppose we have pairs of training samples
$\{(\bi{x}_i,\bi{y}_i)\}_{i=1}^N$.
$a$ and $b$ can be found by solving the Lagrangean
\begin{eqnarray}
  {\cal L}_0 &=& {\rm E}[(u - {\rm E}[u])(v - {\rm E}[v])] \nonumber \\
  && -{\lambda_1\over2}{\rm E}[(u-{\rm E}[u])^2] \nonumber \\
  && -  {\lambda_2\over2}{\rm E}[(v-{\rm E}[v])^2].
\end{eqnarray}
However, the Lagrangean is ill-posed as it is when
the dimensionalities of the Hilbert spaces are large.
Therefore, we introduce a quadratic regularization term and
we get well-posed Lagrangean,
\begin{equation}
  {\cal L} = {\cal L}_0 + {\eta\over2} (\|a\|^2 + \|b\|^2),
\end{equation}
where $\eta$ is a regularization constant.
Note that the average of $u$ is given by
\begin{equation}
  {\rm E}[u] = {1\over N}\sum_{i} \inner{a}{\phi_x(\bi{x}_i)},
\end{equation}
and the average of $uv$ is given by
\begin{equation}
  {\rm E}[uv] = {1\over N}\sum_{i,j} \inner{a}{\phi_x(\bi{x}_i)}
  \inner{b}{\phi_y(\bi{y}_i)}.
\end{equation}
Now, from the condition that the derivative of
${\cal L}$ by $a$ is equal to 0, we get
\begin{equation}
  a = \sum_i \alpha_i \phi_x(\bi{x}_i),
\end{equation}
where $\alpha_i$ is a schalar, then as a result, we have
\begin{equation}
  u = \sum_i \alpha_i \inner{\phi_x(\bi{x}_i)}{\phi_x(\bi{x})}.
\end{equation}
Therefore, $u$ can be calculated by only inner products in $H_x$.
Kernel trick used in SVM uses a kernel function $k_x(\bi{x}_1, \bi{x}_2)$
instead of the inner product between $\phi_x(\bi{x}_1)$ and $\phi_x(\bi{x}_2)$.
In practice, since we don't need an explicit form of $\phi_x$,
we first determine $k_x$ that can be decomposed in the form of inner product.
From Mercer theorem, the symmetric positive definite kernel $k_x$
can be decomposed into the inner product form.

Let us rewrite ${\cal L}$ by the kernel.
First, let $\bi{\alpha}=(\alpha_1,$ $\ldots,$ $\alpha_N)^{\rm T}$,
$\bi{\beta}=(\beta_1,\ldots,\beta_N)^{\rm T}$, and we define
the matrices
\begin{eqnarray}
  (K_x)_{ij} = k_x(\bi{x}_i, \bi{x}_j),
\end{eqnarray}
\begin{eqnarray}
  (K_y)_{ij} = k_y(\bi{y}_i, \bi{y}_j).
\end{eqnarray}
Then, we obtain ${\cal L}$ by
\begin{eqnarray}
  {\cal L} &=& \bi{\alpha}^{\rm T} M \bi{\beta} \nonumber\\
  && - {\lambda_1\over2}\bi{\alpha}^{\rm T} L \bi{\alpha} 
   - {\lambda_2\over2}\bi{\beta}^{\rm T} N \bi{\beta} 
\end{eqnarray}
where
\begin{eqnarray}
  M &=& {1\over N} K_x^{\rm T} J K_y, \\
  L &=& {1\over N} K_x^{\rm T} J K_x + \eta_1 K_x, \\
  N &=& {1\over N} K_y^{\rm T} J K_y + \eta_2 K_y, \\
  J &=& I - {1\over N}{\bf 1}{\bf 1}^{\rm T}, \\
  {\bf 1} &=& (1,\ldots,1)^{\rm T},
\end{eqnarray}
and $\eta_1 = \eta / \lambda_1$, $\eta_2 = \eta / \lambda_2$.

If $\eta > 0$ is satisfied,
$L$ and $N$ are positive definite almost surely, and
we can show $\lambda_1=\lambda_2=\lambda$ from the constraint,
then as a result we have a generalized eigenvalue problem
for $\bi{\alpha}$, $\bi{\beta}$
\begin{eqnarray}
  M \bi{\beta} &=& \lambda L \bi{\alpha}, \\
  M^{\rm T} \bi{\alpha} &=& \lambda N \bi{\beta}, 
\end{eqnarray}

It can be solved by generalized eigenvalue problem package
or Cholesky decomposition of $L$ and $M$.

\section{Computer simulation}

\subsection{Simulation 1}

We generate training samples and test samples independently
as follows.
First $\theta$ is generated from the uniform distribution on 
$[-\pi,\pi]$, and then a pair of two dimensional variables
$\bi{x}$ and $\bi{y}$ are generated by
\begin{equation}
  \bi{x} =\left( \begin{array}{c}
    \theta\\
    \sin3\theta
  \end{array} \right)+ \bi{\epsilon}_1,
\end{equation}
\begin{equation}
  \bi{y} = {\rm e}^{\theta/4}\left(
  \begin{array}{c}
    \cos2\theta\\
    \sin2\theta
  \end{array} \right) + \bi{\epsilon}_2,
\end{equation}
where $\bi{\epsilon}_1$, $\bi{\epsilon}_2$ are independent two
dimensional Gaussian noise with a standard deviation 0.05.

\begin{figure}[ht]
\begin{center}
  \includegraphics[width=.4\textwidth]{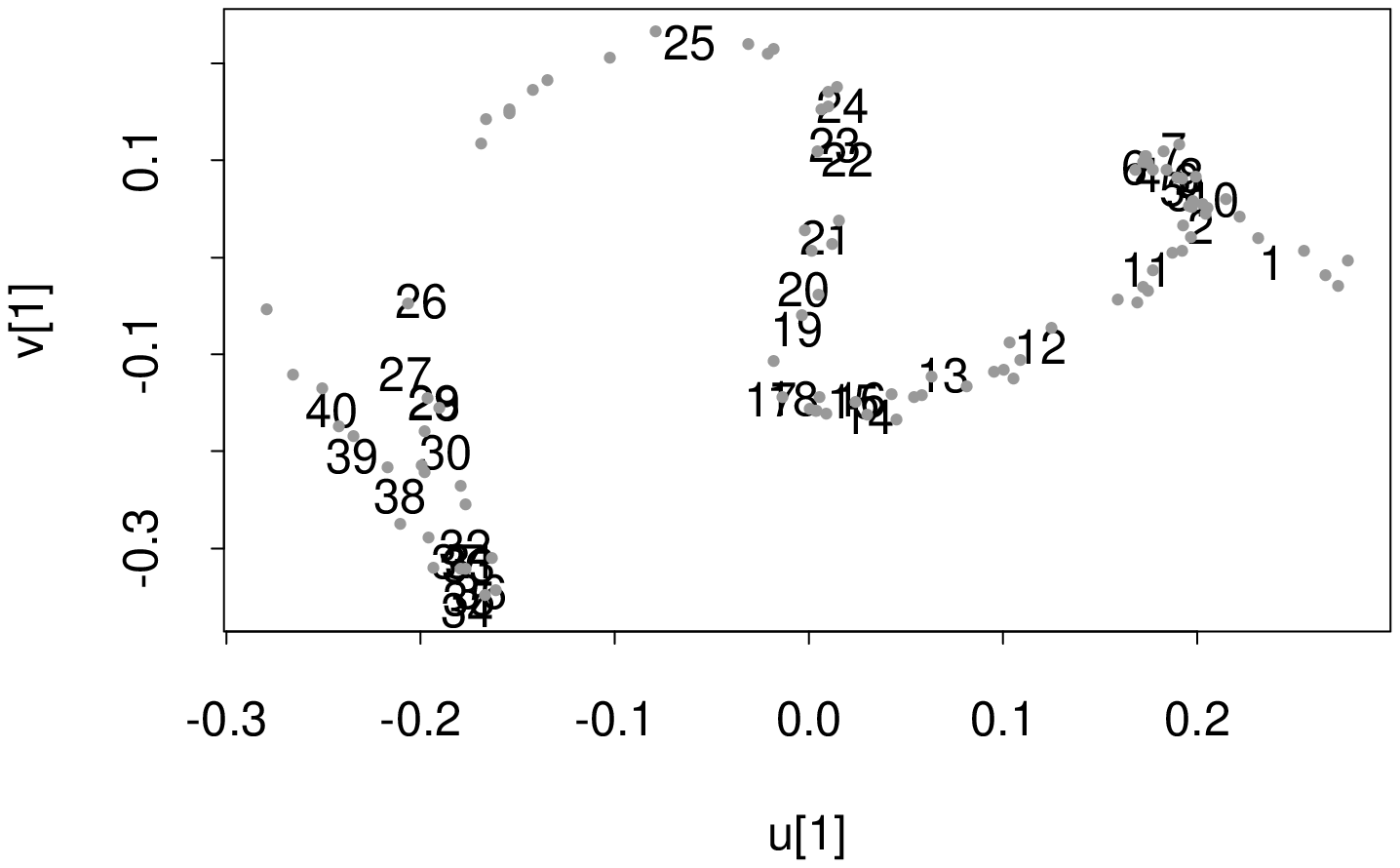}  \\
  \includegraphics[width=.4\textwidth]{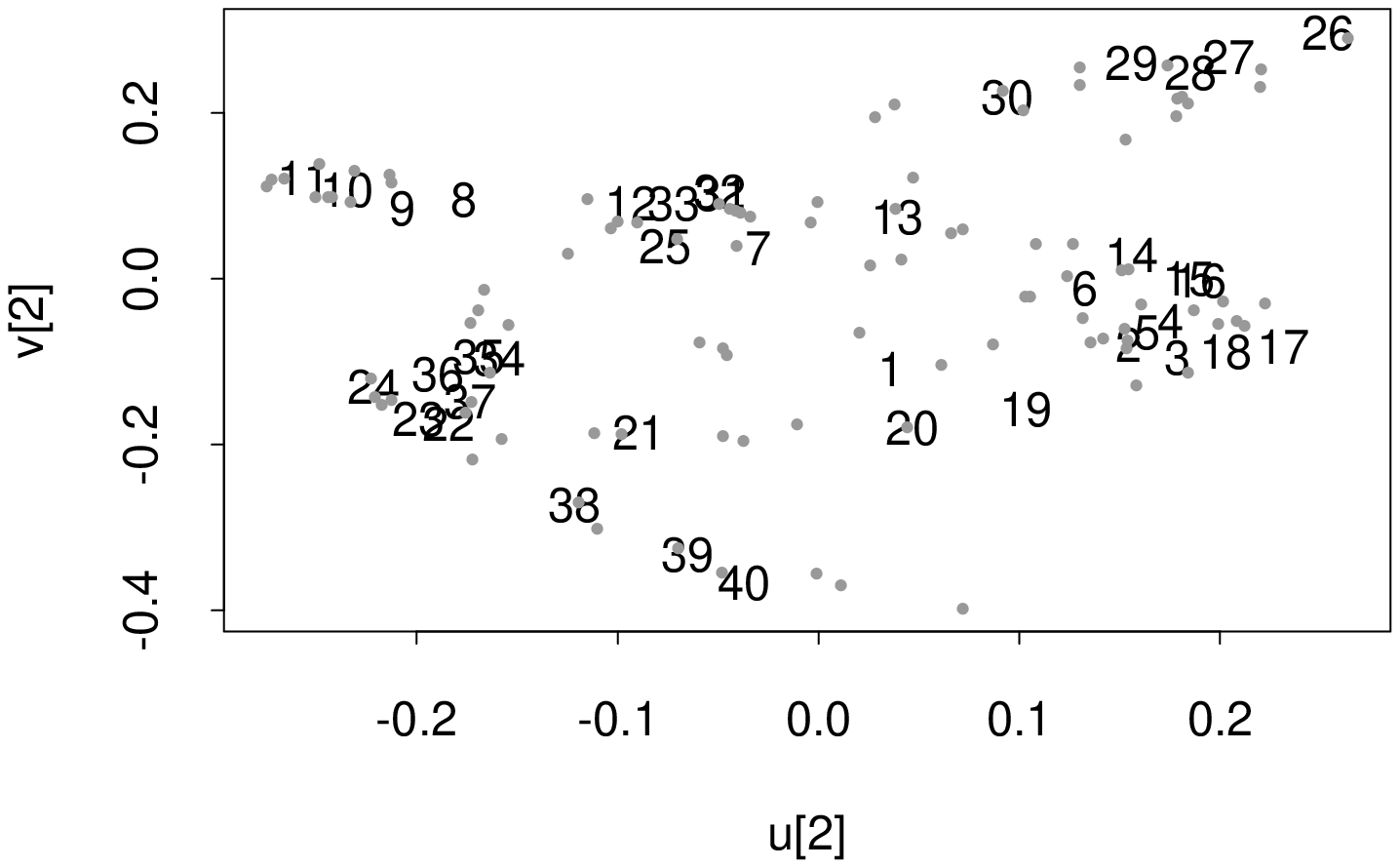}  
  \caption{Simulation 1. $x$--$y$ plot for CCA.
 The numbers represent the increasing order of training samples for $\theta$}
\label{fig:test1-cca-xy}
\end{center}
\end{figure}

\begin{figure}[ht]
\begin{center}
  \includegraphics[width=.4\textwidth]{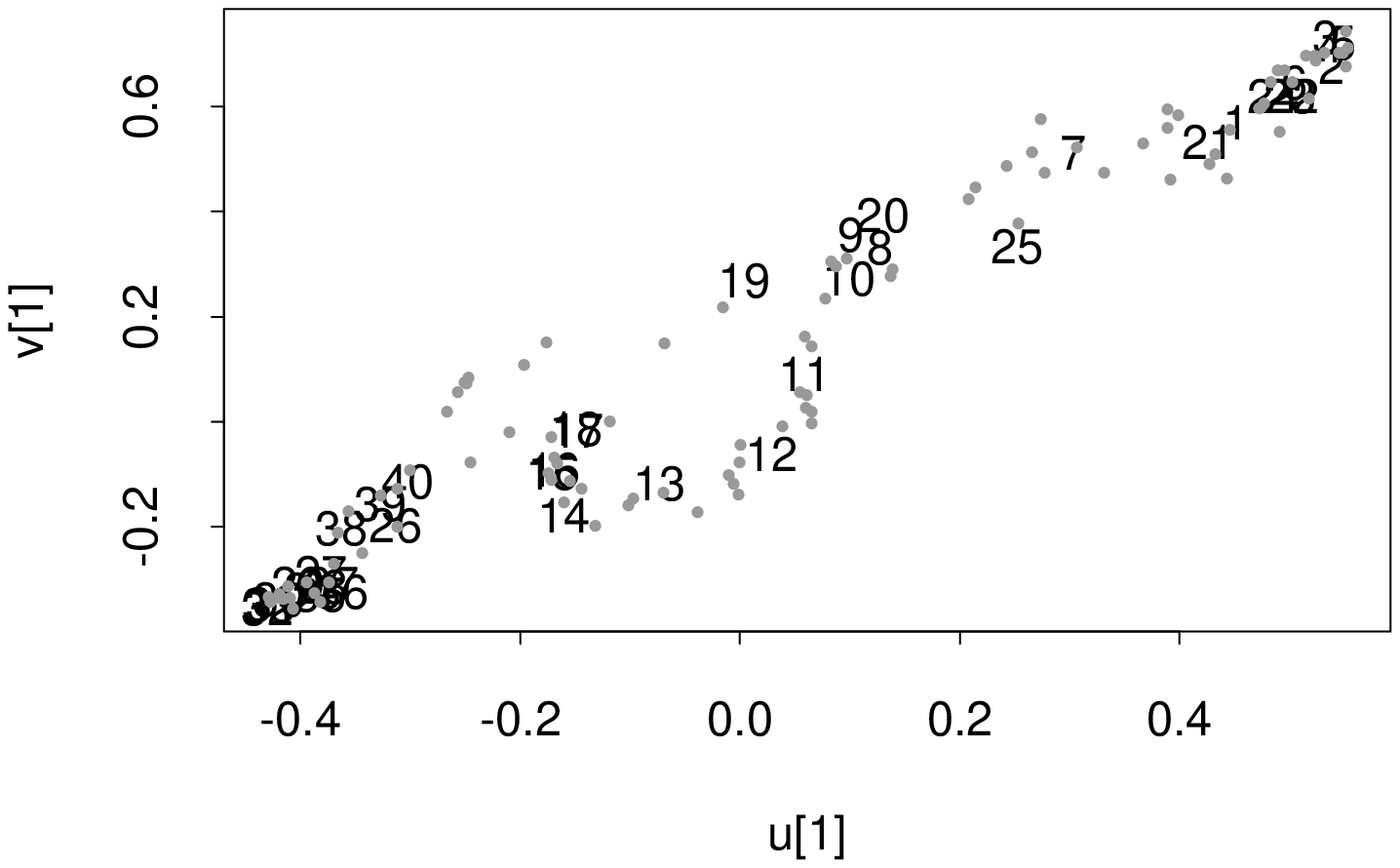}  \\
  \includegraphics[width=.4\textwidth]{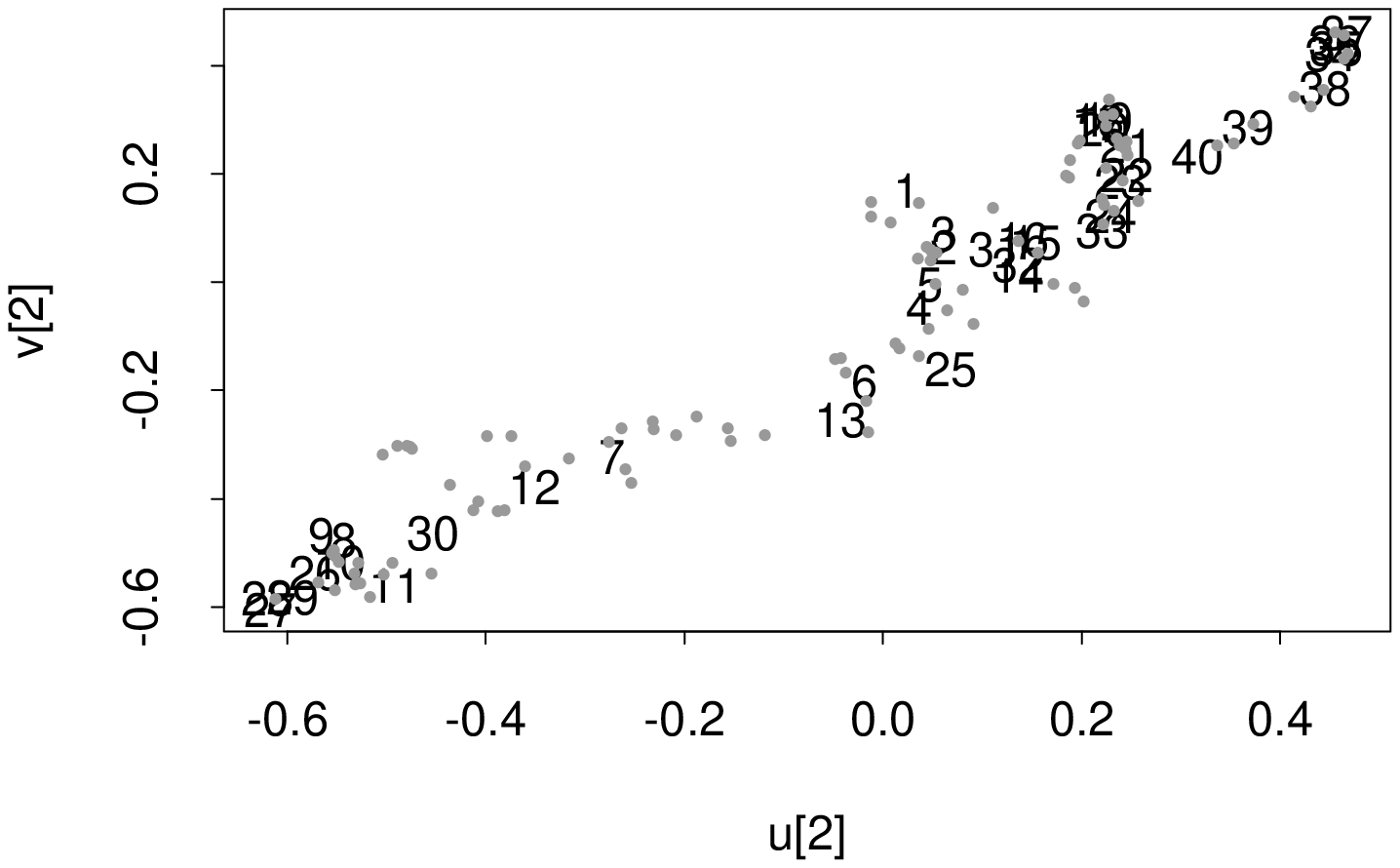}  
  \caption{Simulation 1. $x$--$y$ plot for kernel CCA}
\label{fig:test1-kcca-xy}
\end{center}
\end{figure}

We test for 40 training samples and 100 test samples.
The $x$--$y$ scatter plot of (linear) CCA is shown 
fig.\ref{fig:test1-cca-xy}.
The correlation coefficients are as follows, where
the values for test samples are in the braces.
\begin{center}
\begin{tabular}[c]{|c|rr|rr|}
\hline
& \multicolumn{2}{c|}{$v_1$} & \multicolumn{2}{c|}{$v_2$} \\ \hline
$u_1$ & 0.71 & (0.40) & 0.00 & (0.09) \\ \hline
$u_2$ & 0.00 & (0.00) & 0.27 & (0.19) \\ \hline
\end{tabular}
\end{center}
The $x$--$y$ plot of kernel CCA is shown in fig.\ref{fig:test1-kcca-xy}.
We used Gaussian kernel 
\begin{equation}
  k(\bi{x}_1,\bi{x}_2) = \exp(-{\|\bi{x}_1-\bi{x}_2\|^2\over2\sigma^2}),
\end{equation}
both for $\bi{x}$ and $\bi{y}$,
where parameters are take by $\eta=1.0$, $\sigma=1.0$.
The correlation coefficients are as follows,
where the values for test samples are in the braces.
\begin{center}
\begin{tabular}[c]{|c|rr|rr|}
\hline
& \multicolumn{2}{c|}{$v_1$} & \multicolumn{2}{c|}{$v_2$} \\ \hline
$u_1$ & 0.98 & (0.95) & 0.00 & (0.02) \\ \hline
$u_2$ & 0.00 & (0.02) & 0.97 & (0.93) \\ \hline
\end{tabular}
\end{center}
We only show upto the second components, though
we have higher components in the kernel CCA.

\subsection{Simulation 2}
This section examines an artificial pattern recognition tasks
in multimodal setting described in the beginning of the paper.

Training samples $\bi{x}$ and $\bi{y}$ are generated randomly from the
uniform distribution on $[0,1]^2$ and make random pairs of training samples.
Each pair of training samples represent a class center.
Test samples are generated by adding an independent Gaussian noise
with standard deviation 0.05 to training samples randomly chosen.

We test 10 training samples (classes) and 100 test samples.

\begin{figure}[ht]
\begin{center}
  \includegraphics[width=.4\textwidth]{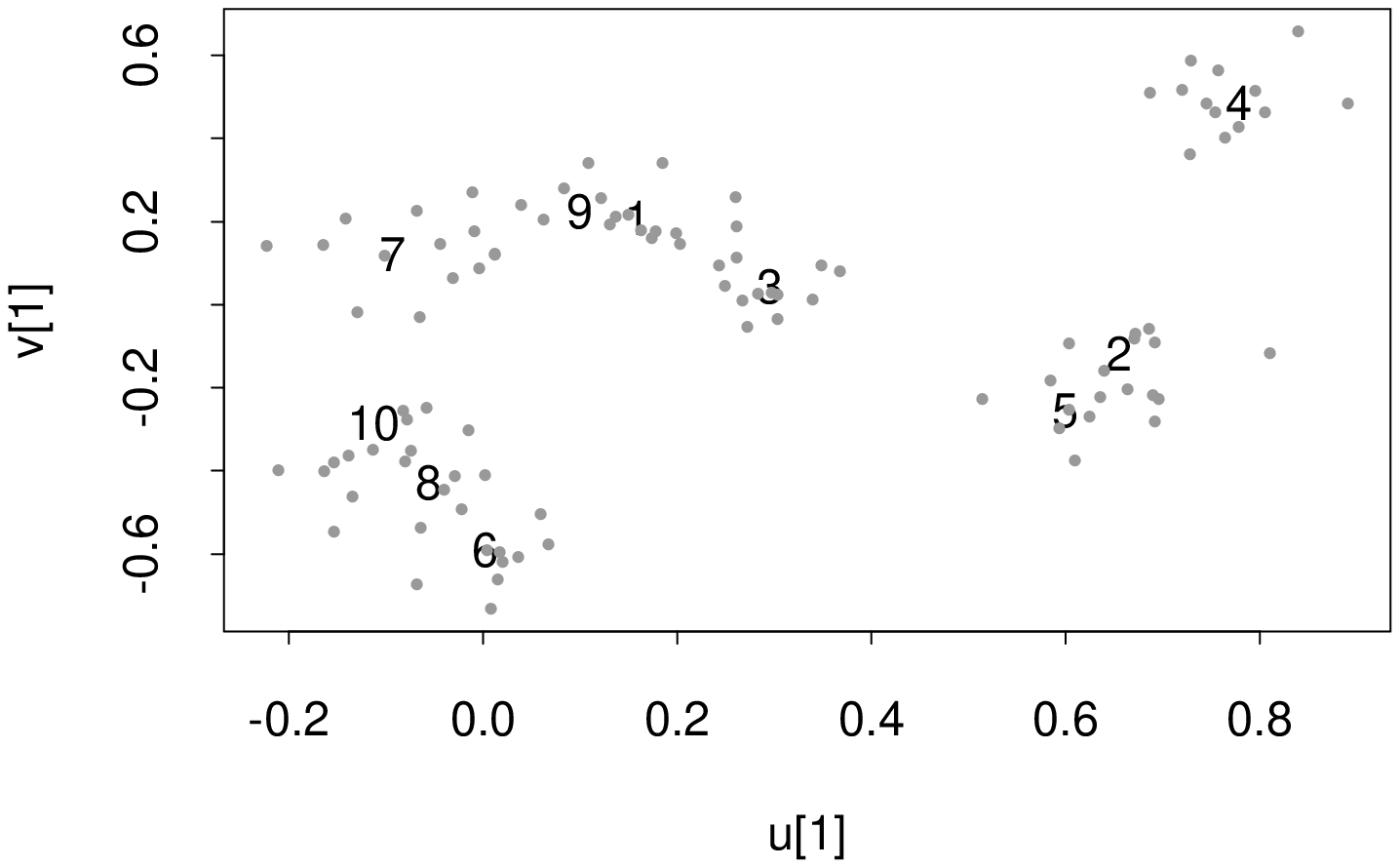}  \\
  \includegraphics[width=.4\textwidth]{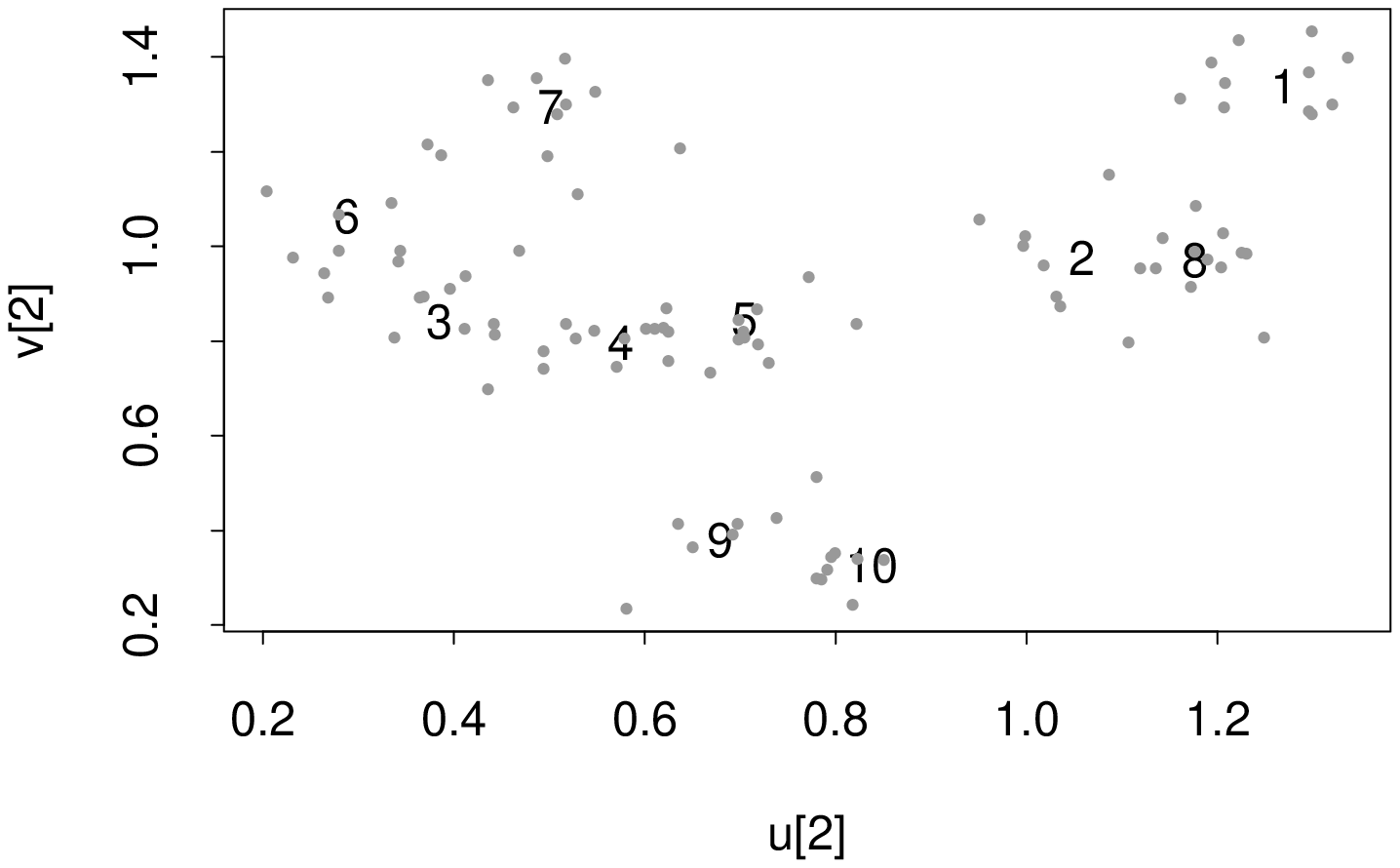}  
  \caption{Simulation 2. $x$--$y$ plot of CCA. The numbers represents
 class centers}
\label{fig:test2-cca-xy}
\end{center}
\end{figure}

$x$--$y$ plot of CCA result is shown in fig.\ref{fig:test2-cca-xy}.
The correlation coefficient between features are as follows,
where the values for test samples are in the braces.

\begin{center}
\begin{tabular}[c]{|c|rr|rr|}
\hline
& \multicolumn{2}{c|}{$v_1$} & \multicolumn{2}{c|}{$v_2$} \\ \hline
$u_1$ & 0.40 & (0.44) & 0.00 & ($-$0.10) \\ \hline
$u_2$ & 0.00 & ($-$0.05) & 0.13 & (0.19) \\ \hline
\end{tabular}
\end{center}

\begin{figure}[ht]
\begin{center}
  \includegraphics[width=.4\textwidth]{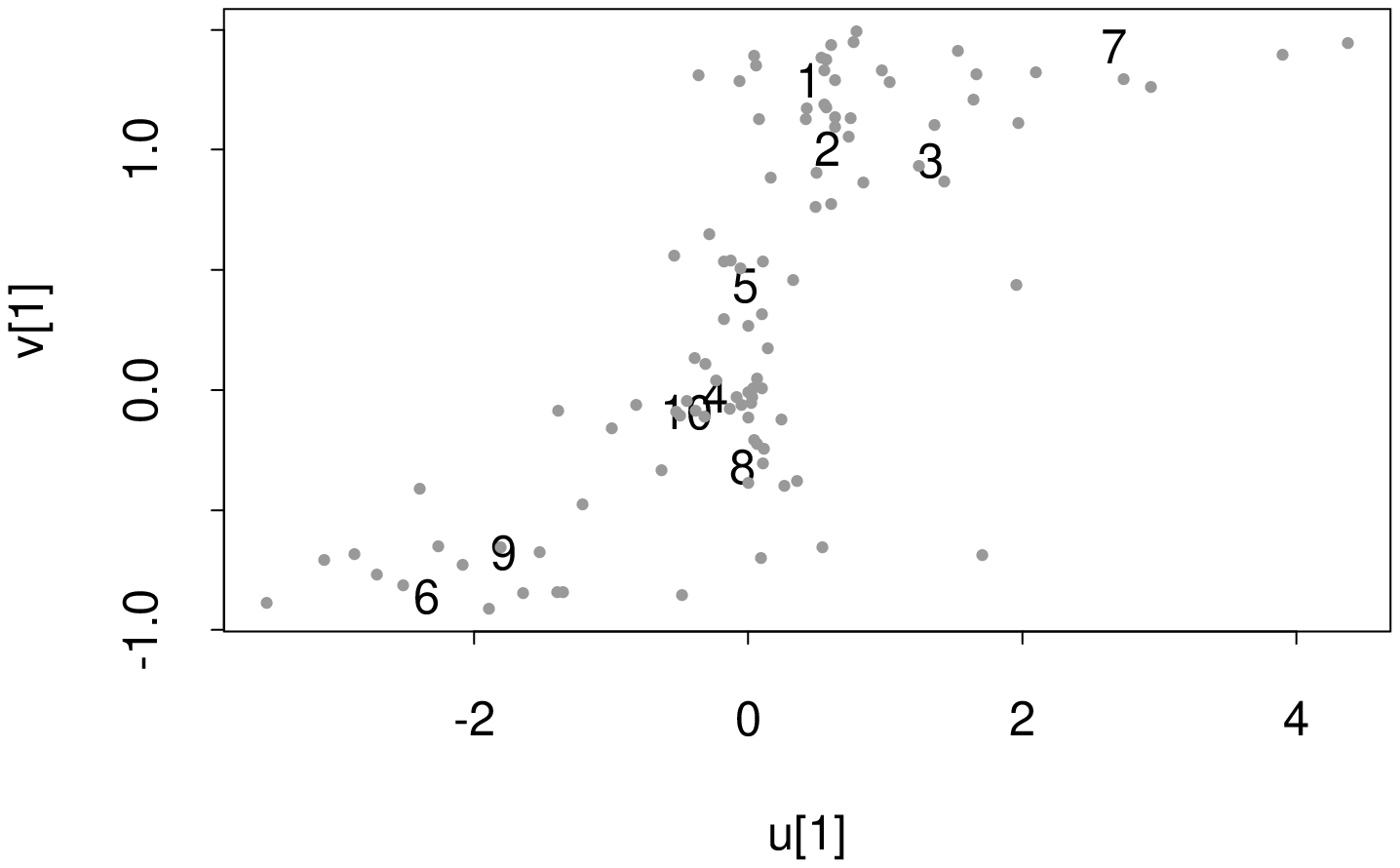}  \\
  \includegraphics[width=.4\textwidth]{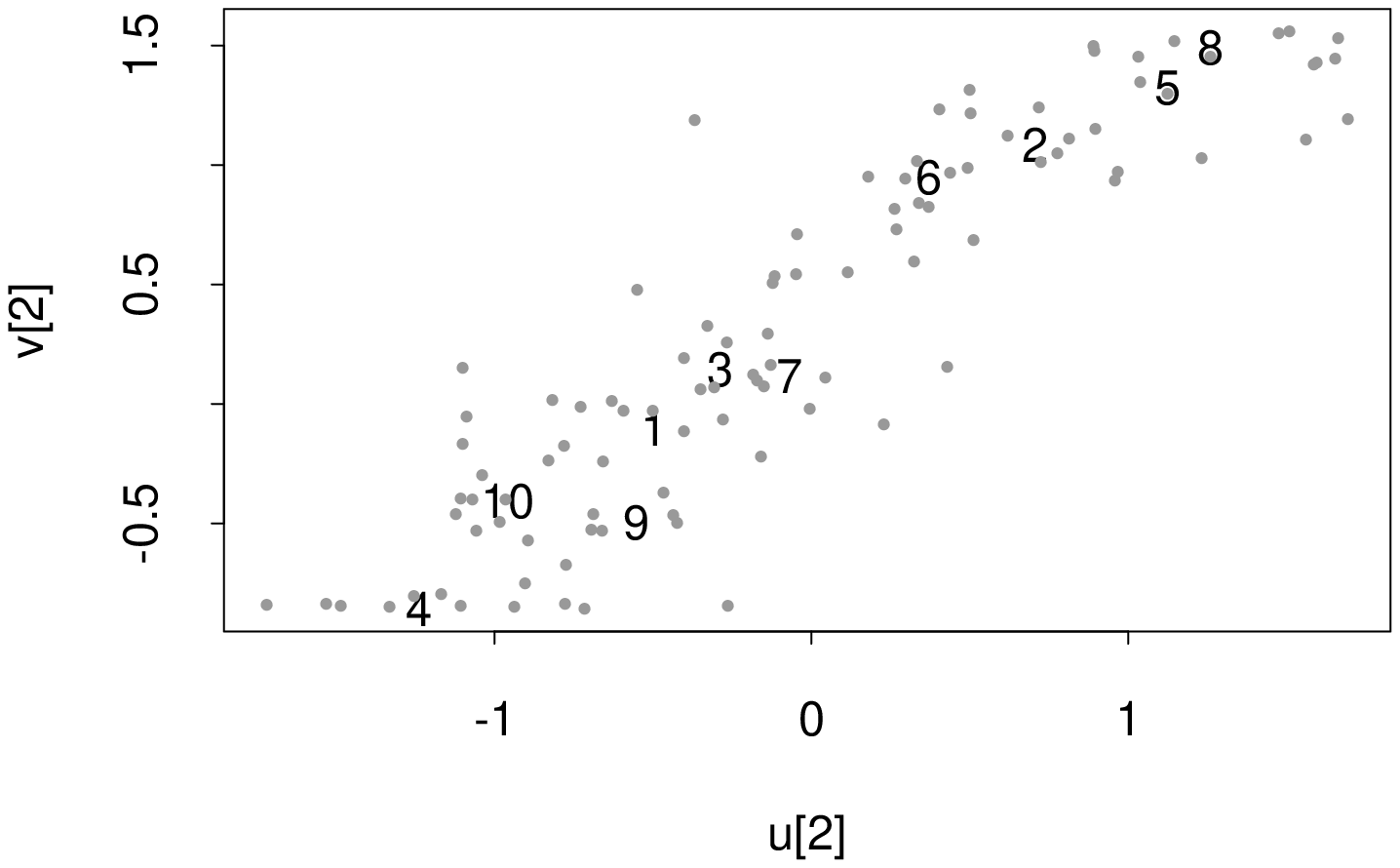}  
  \caption{Simulation 2. $x$--$y$ plot of KCCA}
\label{fig:test2-kcca-xy}
\end{center}
\end{figure}

$x$--$y$ plot of kernel CCA result for the same dataset is shown in 
fig.\ref{fig:test2-kcca-xy}.

We use Gaussian kernel in which parameters are taken
$\eta=0.1$, $\sigma=0.1$.
The correlation coefficients between features are as follows,
where the values of test samples are in braces.

\begin{center}
\begin{tabular}[c]{|c|rr|rr|}
\hline
& \multicolumn{2}{c|}{$v_1$} & \multicolumn{2}{c|}{$v_2$} \\ \hline
$u_1$ & 0.97 & (0.90) & 0.00 & (0.04) \\ \hline
$u_2$ & 0.00 & (0.01) & 0.95 & (0.88) \\ \hline
\end{tabular}
\end{center}

\section{Concluding remarks}

\subsection{Kernel method and regularization}

We have proposed kernel canonial correlation analysis in which
the kernel method is incorporated in the kernel method.
It is similar to SVM that the point is nonlinearization by
kernel method and avoiding overfitting by regularization technique.

In general, it is important to determine the regularizaion parameter.
Moreover, the selection of kernel form is crucial for the performance.
Although all parameters are determined by
hand in the simulations of this paper, we can take more systematic
approaches, such as resampling methods like cross-validation
and emprical Bayes approaches\cite{tipping}.
In such techniques, we usually need iterative algorithms which
is time consuming and is also likely to be trapped into a local optimum.
To examine such issues are future work.

As for regularization term,
we can use $\|\bi{\alpha}\|^2$ and $\|\bi{\beta}\|^2$ instead
of the quadratic term of regularization in this paper.
In the kernel discriminant analysis described below, such a different
type of regularization term is used.
The time complexities for both types are same and
empirically we are not able to find significant difference of performance.
However, we may need more realistic experiments.

\subsection{Relation to kernel discriminant analysis}

The canonical correlation analysis is closely related to the
Fisher's discriminant analysis (FDA), which finds a mapping that minimizes
the inner-class variance as well as maximizes inter-class variance
for effective pattern recognition.
FDA can be considered as a special case of CCA.
Mika et al\cite{mika} has proposed a kernel method for FDA, which
is not strictly included into the kernel CCA because
the kernel FDA does not transform the class label by nonlinear mapping.
For both in kernel CCA and kernel FDA, it is difficult to obtain
sparse representation of mapping.
It would be promising idea to incorporate the sparsity as a utility function.

\subsection{Future issues from the information theory}

The author's group has been proposed the multimodal independent
component analysis (multimodal ICA) which extends the CCA by incorporating the
information theoretic viewpoint\cite{akaho}.
The transformation is restricted to linear and it has been sometimes
difficult to extract useful features from nonlinearly related multivariates.
Now we can raise a question: Can we integrate the kernel CCA
with multimodal ICA in order to extract useful features?

The answer for this question depends on the property of given data.
If the noise level is low as in the simulation of this paper,
the regularization constants are set to small values and
it is desired that the correlation coefficients are almost 1.
We cannot expect the performance is improved by multimodal ICA because
the correlation coefficient close to 1 already achieves
a large amount of mutual information.

On the other hand, when the noise level is large,
the multimodal ICA possibly improves the performance.
However, in such a case, the linear CCA is sometimes enough in practice.
If we learn a multiple value function as in the aquisition of
multiple consept\cite{akaho2}, it may worth trying because
the correlation coefficients are small even if
the noise level is low.

Let us consider further the case the noise level is low.
From the result of the simulations in the previous section,
samples are mapped into a few
clusters that will make regression between $\bi{x}$ and $\bi{y}$ difficult.
In such a case, the distribution of $\bi{u}$ and $\bi{v}$ is desired
to be scattered.
From the information theoretic viewpoint,
the feature space is preferrable to have large amount of entropy.
Since the distribution with largest entropy is Gaussian
under fixed average and variance, the Gaussianity can be used
for the utility function.
For example, the third and forth cumulants are preferrable to be
as small as possible.
It seems opposite from the projection pursuit and independent component
analysis, but it may be caused from the difference of the purpose
that ICA is for visualization while our task is for regression.
The assumption of noise is also different.
These issues are related to the sparsity stated in the previous section,
and it is also future work.

\end{document}